\documentclass[11pt]{article}

\usepackage[preprint]{acl}

\usepackage{times}
\usepackage{latexsym}
\usepackage{fvextra}

\usepackage{placeins}

\usepackage[T1]{fontenc}
\usepackage[utf8]{inputenc}

\usepackage{microtype}

\usepackage{inconsolata}

\usepackage{graphicx}

\title{Distilling Literary Sentiment from LLM Annotations into an Ordinal BERT Model}

\author{Artem Suslov \\ Hokkai-Gakuen University, Sapporo, Japan
         \And
         Takehiro Hashimoto \\ Chuo University, Tokyo, Japan }

\begin{document}
\maketitle
\begin{abstract}
The scarcity of human-annotated literary corpora limits the
development of domain-specific sentiment analysis models.
This study investigates whether LLM-generated annotations
can supervise smaller models for literary sentiment analysis.
We fine-tune five BERT models using ordinal regression on
10,889 literary sentences annotated by Gemma 4 31B on a
ten-point sentiment scale. The models are evaluated against
1,923 human-annotated sentences from Hemingway's
\textit{The Old Man and the Sea}, alongside the teacher model
and conventional baselines.

BERT-large achieves a Kendall's $\tau$ of 0.589, approaching
the teacher's 0.604. Although a constant-neutral predictor
achieves the lowest overall mean absolute error (MAE),
BERT-base outperforms the teacher on non-neutral sentences,
achieving an MAE of 0.726 versus 1.116.
These findings demonstrate the potential of LLM-generated
supervision for developing smaller literary sentiment models,
while highlighting the limitations of aggregate error metrics
on predominantly neutral datasets.

\end{abstract}

\section{Introduction}
Sentiment analysis (SA) has become an established method in Digital Humanities (DH), particularly for investigating emotional patterns and narrative sentiment arcs in literary texts \citep{ohman_emotionarcs_2024}.
However, most available sentiment classifiers were trained on non-literary domains, such as product reviews and social media \citep{barbieri-etal-2020-tweeteval}, while human-annotated literary datasets remain scarce. Consequently, their application to literary texts presents challenges arising from implicit and context-dependent emotional expressions
\citep{bizzoni_sentiment_2024,feldkamp_comparing_2024}.

Another limitation concerns sentiment representation. Most existing classifiers predict nominal categories, such as negative, neutral, and positive, which are often converted into continuous valence scores for literary analysis \citep{bizzoni_sentiment_2024}. Ordinal regression provides an alternative that explicitly models the ordering of sentiment categories
\citep{cao_rank_2020}. Meanwhile, LLMs offer opportunities for generating fine-grained sentiment annotations, although their repeated use for inference across large literary corpora can be computationally demanding \citep{ashir_bert_2024}.

This study investigates whether LLM-generated annotations can provide effective supervision for smaller, domain-specific sentiment models. Specifically, we address two research questions: 

\begin{itemize}
  \item (RQ1) Can BERT models trained on LLM-generated ordinal annotations generalize to independently human-annotated literary text?
  \item (RQ2) How does predictive performance vary across different BERT configurations and sentiment ranges?
\end{itemize}

To address these questions, we make three contributions: (1) we construct a corpus of 10,889 literary sentences annotated by a 31B-parameter LLM on a ten-point sentiment scale; (2) we fine-tune five BERT models using a shared ordinal regression objective; and (3) we evaluate their generalization against independent human annotations, comparing their performance with the teacher model, general-domain classifiers, and a constant-neutral baseline. Our results demonstrate that the fine-tuned models approach the teacher's rank correlation with human judgments while achieving lower prediction errors on non-neutral literary sentences.

\section{Related Work}
SA has become an established method in computational literary studies, particularly for examining narrative sentiment arcs and emotional patterns in literary texts \citep{ohman_emotionarcs_2024}. Transformer-based models, including BERT and RoBERTa, have been increasingly employed for these tasks. However, their effectiveness remains constrained by the complexity of literary language, where sentiment is frequently implicit, contextual, or conveyed through figurative expressions \citep{bizzoni_sentiment_2024,feldkamp_comparing_2024}.

Recent advances in LLMs have enabled their use both as sentiment
classifiers \citep{krugmann_2024,zhang-etal-2024-sentiment,miah_multimodal_2024,wang2024chatgptgoodsentimentanalyzer,coutinho_leveraging_2026}
and as annotators for constructing training datasets
\citep{hellwig-etal-2026-llm,huang-simpson-2026-efficient,zhang-etal-2025-targeted,xie-etal-2025-comprehensive,shvets-2025-emo,ashir_bert_2024,pangakis-wolken-2024-knowledge}. Previous experiments demonstrated
that LLMs with approximately 8B parameters or more can achieve
rank correlations comparable to or higher than conventional
Transformer sentiment classifiers on literary texts
\citep{suslov_tracebale_2026}. However, direct LLM inference
is substantially more computationally demanding than using
smaller encoder-based models.

In DH, custom SA models are being predominantly fine-tuned through nominal classification. A recently developed fiction-oriented RoBERTa model converts continuous sentiment annotations into three nominal classes and reports evaluations across several literary datasets \citep{pionzola_sentiment-fiction_nodate}. Ordinal regression offers an alternative by explicitly preserving the ordering of sentiment categories \citep{cao_rank_2020, biasetton_how_2026}. \citeauthor{biasetton_how_2026} demonstrated its application to BERT-based sentiment prediction using a five-point scale, although outside the literary domain. Building on these approaches, we investigate whether LLM-generated annotations can provide effective supervision for a ten-level ordinal BERT model trained specifically for literary SA and evaluated against independent human judgments.

\section{Data: LLM-Annotated Literary Sentiment}
\label{sec:data}

The training corpus consists of 10,889 sentences extracted from literary works in the \citeauthor{projectgutenberg} collection. To ensure diversity across literary works, the extraction procedure randomly sampled one sentence per novel. To reduce the inclusion of paratextual material, such as titles, tables of contents, and section headings, we excluded very short sentences and structural entries (e.g., ``Chapter X'', ``Section X'', and ``Part X''), as well as the opening and closing sentences of each work.

The resulting corpus was annotated using Gemma 4 31B \citep{gemmateam2026gemma4}, which assigned each sentence a sentiment score on a ten-point scale, ranging from 1 (strongly negative) to 10 (strongly positive), with scores of 4--6 representing the near-neutral range. The annotation prompt, generation parameters, and post-processing procedures are described in Appendix~\ref{app:annotation}.

For independent evaluation, we employed the human-annotated corpus of Ernest Hemingway's \textit{The Old Man and the Sea}, introduced by \citet{bizzoni_sentiment_2024}. This dataset comprises 1,923 sentences evaluated by two human annotators on a comparable ten-point valence scale. The evaluation corpus was excluded from model training.

\section{Method: Ordinal BERT}
\label{sec:method}

We fine-tuned five BERT configurations using an ordinal regression architecture inspired by \citet{mccullagh1980regression,cao_rank_2020,wang_ordinal_2026}. Unlike conventional sentiment classifiers that treat sentiment categories as independent labels, ordinal regression explicitly preserves their inherent ordering. This approach is particularly suitable for literary SA, where valence is expressed on a graded scale rather than through discrete polarities.

Given an input sentence $x$, BERT produces a contextual representation, which is projected onto a scalar latent sentiment score $s(x)$. For $K=10$ sentiment levels, the model learns nine monotonically ordered thresholds $\theta_1 < \cdots < \theta_9$, enforced through softplus parameterization. Each threshold defines the cumulative probability:
\begin{equation}
P(Y>k \mid x) =
\sigma(s(x)-\theta_k),
\quad k=1,\ldots,9,
\end{equation}
where $\sigma$ denotes the sigmoid function. Training labels are converted into cumulative binary targets and optimized using binary cross-entropy loss. The ordered thresholds ensure that predictions remain rank-consistent.

At inference, the expected sentiment score is calculated as:
\begin{equation}
\hat{y} =
1+\sum_{k=1}^{9}P(Y>k\mid x).
\end{equation}
This produces a continuous sentiment estimate on the original 1--10 scale while preserving the ordinal structure of the training labels, without relying on post-hoc transformations of nominal sentiment probabilities.

\section{Experimental Setup}
\label{sec:experiments}

We fine-tuned five pretrained BERT models of different sizes:
Large, Base, Medium, Small, and Mini, described in Appendix~\ref{app:bert_models}. The LLM-annotated corpus
was divided into training ($n=8,711$), validation ($n=1,089$),
and test ($n=1,089$) sets. All models used identical
optimization hyperparameters, detailed in
Appendix~\ref{app:training}.

\paragraph{Baselines.}
We compared the models against Gemma 4 31B (teacher),
two general-domain RoBERTa classifiers
(\texttt{twt\_roberta} and \texttt{twt\_roberta\_xlm}) \citep{barbieri-etal-2020-tweeteval,barbieri2022xlmt},
and a constant-neutral baseline (score = 5). RoBERTa model scores were linearly scaled to 1-10 range.
BERT-tiny also converged to constant predictions
and is reported alongside this baseline.

\paragraph{Evaluation.}
Models were evaluated on 1,923 sentences from Hemingway's
\textit{The Old Man and the Sea}, using the average scores
of two human annotators. This corpus was excluded from
training and validation. We calculated MAE, Kendall's $\tau$, and precision and accuracy within one sentiment level ($\pm 1$). MAE and $\tau$ used original
scores; tolerance-based precision was macro-averaged across
observed predicted values.

To account for the predominance of neutral scores, we additionally evaluated 305 non-neutral sentences, defined by human scores below 4 or above 6.

\section{Results}
\label{sec:results}
\subsection{Non-Literary Domain SA}
Table~\ref{tab:model_comparison} presents the performance of the fine-tuned BERT models, the teacher LLM, general-domain RoBERTa classifiers, and the constant-neutral baseline against human sentiment annotations \footnote{Training dataset, training code and released models are available on \url{https://github.com/susurofu/literary-bert-distilling}}.

\begin{table*}[!t]
\centering
\small

\resizebox{\textwidth}{!}{
\begin{tabular}{lccccccc}
\hline

& \multicolumn{4}{c}{\textbf{All sentences ($n=1,923$)}}
& \multicolumn{3}{c}{\textbf{Non-neutral sentences ($n=305$)}} \\

\cline{2-5}
\cline{6-8}

\textbf{Model}
& \textbf{MAE}
& \textbf{Kendall's $\tau$}
& \textbf{Prec. ($\pm$1)}
& \textbf{Acc. ($\pm$1)}
& \textbf{MAE}
& \textbf{Prec. ($\pm$1)}
& \textbf{Acc. ($\pm$1)} \\

\hline

Gemma 4 31B (teacher)
& 0.822 & \textbf{0.604} & 0.460 & 0.741
& 1.116 & 0.522 & 0.600 \\

twt\_roberta
& 0.972 & 0.497 & 0.176 & 0.695
& 1.656 & 0.306 & 0.223 \\

twt\_roberta\_xlm
& 1.261 & 0.505 & 0.125 & 0.530
& 1.613 & 0.301 & 0.308 \\

\hline

BERT-large
& 0.769 & 0.589 & 0.614 & 0.787
& 0.887 & 0.571 & 0.711 \\

BERT-base
& 0.733 & 0.580 & 0.723 & 0.815
& \textbf{0.726} & \textbf{0.698} & \textbf{0.803} \\

BERT-medium
& 0.752 & 0.489 & 0.659 & 0.808
& 0.998 & 0.669 & 0.675 \\

BERT-small
& 0.717 & 0.468 & 0.708 & 0.837
& 0.982 & 0.668 & 0.698 \\

BERT-mini
& 0.766 & 0.402 & 0.732 & 0.829
& 0.969 & 0.692 & 0.708 \\

\hline

BERT-tiny (constant-neutral)
& \textbf{0.597} & --
& \textbf{0.841} & \textbf{0.841}
& 1.962 & 0.000 & 0.000 \\

\hline

\end{tabular}
}

\caption{
Performance against human annotations (n = 1,923) and on
non-neutral sentences (n = 305; scores <4 or >6).
MAE denotes mean absolute error; precision and accuracy
use a tolerance of $\pm 1$. Bold indicates the best
result per metric. Kendall's $\tau$ is undefined for
constant predictions.
}

\label{tab:model_comparison}

\end{table*}

\paragraph{Overall performance.}
All five fine-tuned BERT models achieved lower MAE than the general-domain RoBERTa baselines. Among the student models, BERT-small obtained the lowest MAE (0.717) and the highest accuracy within one sentiment level (0.837).

However, none of the evaluated models outperformed the constant-neutral baseline, which achieved an MAE of 0.597 and an accuracy of 0.841. This result reflects the predominance of near-neutral sentiment in the evaluation corpus, where 84.1\% of human reference scores fell within the interval from 4 to 6.

\paragraph{Performance on non-neutral sentences.}
To examine predictive performance beyond the dominant neutral range, we separately evaluated 305 sentences with human sentiment scores below 4 or above 6.

All five fine-tuned BERT models outperformed both the teacher LLM and the general-domain RoBERTa classifiers in MAE and accuracy within one sentiment level on this subset.

BERT-base achieved the strongest performance, with an MAE of 0.726 and an accuracy of 0.803, compared with 1.116 and 0.600 for the teacher model, respectively. The constant-neutral baseline achieved an MAE of 1.962 and an
zero accuracy on this subset.

These results demonstrate that the performance advantage of the constant-neutral baseline on the complete corpus does not extend to sentences expressing sentiment outside the near-neutral range.

\paragraph{Rank correlation and model size.}
Rank-based evaluation revealed a different pattern.
The teacher achieved the highest Kendall's $\tau$ (0.604),
followed by BERT-large (0.589) and BERT-base (0.580),
both exceeding the general-domain RoBERTa baselines
(0.497 and 0.505). Smaller BERT configurations showed
weaker agreement, with $\tau$ declining from 0.489
for BERT-medium to 0.402 for BERT-mini.

\subsection{Literary Domain SA}

Overall, performance depended on the evaluation metric
and sentiment distribution. BERT-small achieved the lowest
overall MAE among students, BERT-base performed best on
non-neutral sentences, and BERT-large most closely
approached the teacher's rank correlation with human judgments.

\section{Discussion}

\subsection{LLM Supervision and Ordinal Modeling}

Trained on LLM-annotated sentences sampled across different
literary works, the BERT models were evaluated on Hemingway's
stylistically distinctive \textit{The Old Man and the Sea}.
Despite this difference, BERT-large and BERT-base approached
the teacher's rank correlation with human judgments
($\tau=0.589$ and $0.580$ versus $0.604$), while BERT-base
outperformed the teacher on non-neutral MAE (0.726 versus 1.116).
These findings suggest that LLM-generated supervision can
transfer meaningful sentiment information from a stylistically
diverse training corpus to a distinct literary work.

Nevertheless, evaluation on a single novel limits claims
of broader stylistic generalization. Furthermore, the study
did not examine alternative architectures, such as conventional
regression or three-class nominal classification, or assess
training variability across random seeds. Consequently, the
specific contribution of ordinal modeling and the stability
of the results remain to be established.

\subsection{Sentiment Distribution and Evaluation}

The results highlight a common challenge in SA:
imbalanced sentiment distributions can favor models predicting
the dominant neutral category \citep{jongeling_2017}.
In our evaluation corpus, 84.1\% of human annotations
fell within the near-neutral range of 4--6. Consequently,
a constant-neutral predictor achieved the lowest overall MAE,
despite capturing no sentiment variation.

However, identifying non-neutral passages may have greater
heuristic value for literary analysis than accurately predicting
neutral sentences. On the non-neutral subset, all fine-tuned
BERT models outperformed the teacher LLM and general-domain
RoBERTa baselines. These findings suggest that aggregate
error metrics may obscure models' ability to estimate
non-neutral sentiment, limiting their usefulness for
analyzing emotional variation in literary narratives.

One possible improvement involves modifying the training
objective to penalize errors on underrepresented sentiment
categories more heavily. Class-balanced weighting
\citep{cui_2019} and focal loss \citep{lin_2017}
offer established approaches, while distribution-aware
methods address imbalance in continuous prediction
\citep{yang_2021}. Adapting these approaches to ordinal
sentiment modeling remains a direction for future research.

\subsection{Model Size and Computational Efficiency}

Model size influenced predictive performance, with larger BERT
variants generally achieving stronger rank correlations.
BERT-base performed best on non-neutral sentences, while
BERT-large most closely approached the teacher's rank correlation.
Thus, model selection should consider whether absolute sentiment
estimation or relative ordering is more important.

Transferring LLM-generated annotations to smaller encoder models
may also reduce computational requirements for large-scale
literary analysis. However, inference efficiency was not
measured directly, and these potential benefits require
experimental verification.

\section{Conclusion}

This study demonstrates that LLM-generated annotations can
provide effective supervision for literary sentiment analysis,
at least within the examined configuration. Using 10,889
LLM-annotated sentences, we fine-tuned BERT models of
different sizes and evaluated them against independent
human judgments.

The larger models approached the teacher's rank correlation
(BERT-large: $\tau=0.589$; Gemma 4 31B: $0.604$),
while BERT-base outperformed the teacher on non-neutral
sentences (MAE: 0.726 versus 1.116). Even compact BERT
configurations demonstrated meaningful predictive performance,
suggesting that smaller encoders can effectively learn from
LLM-generated annotations.

Nevertheless, the constant-neutral baseline achieved the
lowest overall MAE, highlighting the influence of sentiment
imbalance. Further research should examine alternative
training objectives, prediction calibration, and
generalization across literary genres.

\section*{Limitations}

The model was trained on randomly sampled sentences from Project Gutenberg, annotated using an LLM. No additional filtering was applied to exclude potentially bias-inducing contexts. Consequently, the resulting dataset may inherit biases from the annotating LLM.

In particular, we observed that Gemma 4 31B tended to assign disproportionately negative sentiment scores to sentences describing sensitive topics, such as violence, death, or criminal acts, even when their narrative context conveyed positive or contrasting sentiment. This suggests that the model may associate certain semantic topics with negative valence regardless of their contextual meaning.

This was the only specific annotation bias identified in the present study; however, other systematic biases cannot be excluded.

\section*{Declaration of Generative AI Use}

OpenAI's ChatGPT 5.6 model was used for proofreading and minor stylistic refinements (paraphrasing and summarization under authors' supervision) and assisting with \LaTeX{} layout adjustment. Also, the same model was used for debugging assistance and code review.

% Bibliography entries for the entire Anthology, followed by custom entries
%\bibliography{custom,anthology-overleaf-1,anthology-overleaf-2}

% Custom bibliography entries only
\bibliography{custom}

\begin{thebibliography}{31}
\providecommand{\natexlab}[1]{#1}

\bibitem[{Ashir and Taha(2024)}]{ashir_bert_2024}
Abubakar~M. Ashir and Mohammed~Abdulghani Taha. 2024.
\newblock \href {https://doi.org/10.1177/17248035241291687} {{BERT} with an augmented cross-attention decoder ({BERT}-{ACD}) for binary and fine-grained multiband sentiment detection}.
\newblock \emph{Intelligenza Artificiale: The international journal of the {AIxIA}}, 19(1):3--16.

\bibitem[{Barbieri et~al.(2020)Barbieri, Camacho-Collados, Espinosa~Anke, and Neves}]{barbieri-etal-2020-tweeteval}
Francesco Barbieri, Jose Camacho-Collados, Luis Espinosa~Anke, and Leonardo Neves. 2020.
\newblock \href {https://doi.org/10.18653/v1/2020.findings-emnlp.148} {{T}weet{E}val: Unified benchmark and comparative evaluation for tweet classification}.
\newblock In \emph{Findings of the Association for Computational Linguistics: EMNLP 2020}, pages 1644--1650, Online. Association for Computational Linguistics.

\bibitem[{Barbieri et~al.(2022)Barbieri, Espinosa~Anke, and Camacho-Collados}]{barbieri2022xlmt}
Francesco Barbieri, Luis Espinosa~Anke, and Jose Camacho-Collados. 2022.
\newblock \href {https://aclanthology.org/2022.lrec-1.27/} {{XLM-T}: Multilingual language models in twitter for sentiment analysis and beyond}.
\newblock In \emph{Proceedings of the Thirteenth Language Resources and Evaluation Conference}, pages 258--266. European Language Resources Association.

\bibitem[{Biasetton et~al.(2026)Biasetton, Ricciardi, Salmaso, and Zuccolotto}]{biasetton_how_2026}
Nicolò Biasetton, Riccardo Ricciardi, Luigi Salmaso, and Paola Zuccolotto. 2026.
\newblock \href {https://doi.org/10.1002/asmb.70090} {How well do ratings reflect sentiment? {E}vidence from a large italian review corpus}.
\newblock \emph{Applied Stochastic Models in Business and Industry}, 42(2):e70090.

\bibitem[{Bizzoni and Feldkamp(2024)}]{bizzoni_sentiment_2024}
Yuri Bizzoni and Pascale Feldkamp. 2024.
\newblock \href {https://doi.org/10.46298/jdmdh.13155} {Sentiment analysis for literary texts: Hemingway as a case-study}.
\newblock \emph{Journal of Data Mining \& Digital Humanities}, {NLP}4DH:1--20.

\bibitem[{Cao et~al.(2020)Cao, Mirjalili, and Raschka}]{cao_rank_2020}
Wenzhi Cao, Vahid Mirjalili, and Sebastian Raschka. 2020.
\newblock \href {https://doi.org/10.1016/j.patrec.2020.11.008} {Rank consistent ordinal regression for neural networks with application to age estimation}.
\newblock \emph{Pattern Recognition Letters}, 140:325--331.

\bibitem[{Coutinho et~al.(2026)Coutinho, Braga, Canuto, Pereira, Assunção, Steinmacher, Gerosa, and Garcia}]{coutinho_leveraging_2026}
Daniel Coutinho, Breno Braga, Theo Canuto, Juliana~Alves Pereira, Wesley K.~G. Assunção, Igor Steinmacher, Marco Gerosa, and Alessandro Garcia. 2026.
\newblock \href {https://doi.org/10.1007/s10664-026-10868-6} {Leveraging large language models for sentiment analysis in {GitHub} pull request discussions}.
\newblock \emph{Empirical Software Engineering}, 31(5):140.

\bibitem[{Cui et~al.(2019)Cui, Jia, Lin, Song, and Belongie}]{cui_2019}
Yin Cui, Menglin Jia, Tsung-Yi Lin, Yang Song, and Serge Belongie. 2019.
\newblock \href {https://doi.org/10.1109/CVPR.2019.00949} {{ Class-Balanced Loss Based on Effective Number of Samples }}.
\newblock In \emph{2019 IEEE/CVF Conference on Computer Vision and Pattern Recognition (CVPR)}, pages 9260--9269, Los Alamitos, CA, USA. IEEE Computer Society.

\bibitem[{Devlin et~al.(2019)Devlin, Chang, Lee, and Toutanova}]{devlin-etal-2019-bert}
Jacob Devlin, Ming-Wei Chang, Kenton Lee, and Kristina Toutanova. 2019.
\newblock \href {https://doi.org/10.18653/v1/N19-1423} {{BERT}: Pre-training of deep bidirectional transformers for language understanding}.
\newblock In \emph{Proceedings of the 2019 Conference of the North American Chapter of the Association for Computational Linguistics: Human Language Technologies, Volume 1 (Long and Short Papers)}, pages 4171--4186, Minneapolis, Minnesota. Association for Computational Linguistics.

\bibitem[{Feldkamp et~al.(2024)Feldkamp, Kostkan, Overgaard, Jacobsen, and Bizzoni}]{feldkamp_comparing_2024}
Pascale Feldkamp, Jan Kostkan, Ea~Overgaard, Mia Jacobsen, and Yuri Bizzoni. 2024.
\newblock \href {https://doi.org/10.18653/v1/2024.wassa-1.15} {Comparing tools for sentiment analysis of {D}anish literature from hymns to fairy tales: Low-resource language and domain challenges}.
\newblock In \emph{Proceedings of the 14th Workshop on Computational Approaches to Subjectivity, Sentiment, \& Social Media Analysis}, pages 186--199. Association for Computational Linguistics.

\bibitem[{{Gemma Team}(2026)}]{gemmateam2026gemma4}
{Gemma Team}. 2026.
\newblock \href {https://arxiv.org/abs/2607.02770} {Gemma 4 technical report}.
\newblock \emph{Preprint}, arxiv:2607.02770.

\bibitem[{Hellwig et~al.(2026)Hellwig, Fehle, Kruschwitz, and Wolff}]{hellwig-etal-2026-llm}
Nils~Constantin Hellwig, Jakob Fehle, Udo Kruschwitz, and Christian Wolff. 2026.
\newblock \href {https://doi.org/10.63317/43srcdyc52cd} {{LLM}-as-an-annotator: Training lightweight models with {LLM}-annotated examples for aspect sentiment tuple prediction}.
\newblock In \emph{Proceedings of the Fifteenth Language Resources and Evaluation Conference}, pages 7955--7972, Palma de Mallorca, Spain. ELRA Language Resource Association.

\bibitem[{Huang and Simpson(2026)}]{huang-simpson-2026-efficient}
Wen-Fong~(Xavier) Huang and Edwin Simpson. 2026.
\newblock \href {https://doi.org/10.63317/3b3zxy5qrw8s} {Efficient financial language understanding via distillation with synthetic data}.
\newblock In \emph{Proceedings of the Fifteenth Language Resources and Evaluation Conference}, pages 10242--10254, Palma de Mallorca, Spain. ELRA Language Resource Association.

\bibitem[{Jongeling et~al.(2017)Jongeling, Sarkar, Datta, and Serebrenik}]{jongeling_2017}
Robbert Jongeling, Proshanta Sarkar, Subhajit Datta, and Alexander Serebrenik. 2017.
\newblock \href {https://doi.org/10.1007/s10664-016-9493-x} {On negative results when using sentiment analysis tools for software engineering research}.
\newblock \emph{Empirical Software Engineering}, 22:2543--2584.

\bibitem[{Krugmann and Hartmann(2024)}]{krugmann_2024}
Jan~Ole Krugmann and Jochen Hartmann. 2024.
\newblock \href {https://doi.org/10.1007/s40547-024-00143-4} {Sentiment analysis in the age of generative ai}.
\newblock \emph{Customer Needs and Solutions}, 11(3):1--19.

\bibitem[{Lin et~al.(2020)Lin, Goyal, Girshick, He, and Dollár}]{lin_2017}
Tsung-Yi Lin, Priya Goyal, Ross Girshick, Kaiming He, and Piotr Dollár. 2020.
\newblock \href {https://doi.org/10.1109/TPAMI.2018.2858826} {Focal loss for dense object detection}.
\newblock In \emph{IEEE Transactions on Pattern Analysis and Machine Intelligence}, volume~42, pages 318--327.

\bibitem[{McCullagh(1980)}]{mccullagh1980regression}
Peter McCullagh. 1980.
\newblock \href {https://doi.org/10.1111/j.2517-6161.1980.tb01109.x} {Regression models for ordinal data}.
\newblock \emph{Journal of the Royal Statistical Society: Series B (Methodological)}, 42(2):109--142.

\bibitem[{Miah et~al.(2024)Miah, Kabir, Sarwar, Safran, Alfarhood, and Mridha}]{miah_multimodal_2024}
Md~Saef~Ullah Miah, Md~Mohsin Kabir, Talha~Bin Sarwar, Mejdl Safran, Sultan Alfarhood, and M.~F. Mridha. 2024.
\newblock \href {https://doi.org/10.1038/s41598-024-60210-7} {A multimodal approach to cross-lingual sentiment analysis with ensemble of transformer and {LLM}}.
\newblock \emph{Scientific Reports}, 14(1):9603.

\bibitem[{Pangakis and Wolken(2024)}]{pangakis-wolken-2024-knowledge}
Nicholas Pangakis and Sam Wolken. 2024.
\newblock \href {https://doi.org/10.18653/v1/2024.nlpcss-1.9} {Knowledge distillation in automated annotation: Supervised text classification with {LLM}-generated training labels}.
\newblock In \emph{Proceedings of the Sixth Workshop on Natural Language Processing and Computational Social Science ({NLP}+{CSS} 2024)}, pages 113--131, Mexico City, Mexico. Association for Computational Linguistics.

\bibitem[{Pianzola(2026)}]{pionzola_sentiment-fiction_nodate}
Federico Pianzola. 2026.
\newblock \href {https://huggingface.co/fpianz/sentiment-fiction} {sentiment-fiction}.

\bibitem[{{Project Gutenberg}()}]{projectgutenberg}
{Project Gutenberg}.
\newblock \href {https://www.gutenberg.org/} {Project gutenberg}.

\bibitem[{Shvets(2025)}]{shvets-2025-emo}
Alexander Shvets. 2025.
\newblock \href {https://doi.org/10.18653/v1/2025.findings-acl.10} {Emo pillars: Knowledge distillation to support fine-grained context-aware and context-less emotion classification}.
\newblock In \emph{Findings of the {A}ssociation for {C}omputational {L}inguistics: {ACL} 2025}, pages 174--191, Vienna, Austria. Association for Computational Linguistics.

\bibitem[{Suslov and Hashimoto(2026)}]{suslov_tracebale_2026}
Artem Suslov and Takehiro Hashimoto. 2026.
\newblock Traceable sentiment analysis with large language models out of the box.
\newblock In \emph{Proceedings of {JADH}2026}, volume 2026. {JADH}2026 Program Committee and Local Organizing Committee.

\bibitem[{Turc et~al.(2020)Turc, Chang, Lee, and Toutanova}]{turc2020wellread}
Iulia Turc, Ming-Wei Chang, Kenton Lee, and Kristina Toutanova. 2020.
\newblock \href {https://arxiv.org/abs/1908.08962} {Well-read students learn better: On the importance of pre-training compact models}.
\newblock In \emph{International Conference on Learning Representations}.

\bibitem[{Wang et~al.(2024)Wang, Xie, Feng, Ding, Yang, and Xia}]{wang2024chatgptgoodsentimentanalyzer}
Zengzhi Wang, Qiming Xie, Yi~Feng, Zixiang Ding, Zinong Yang, and Rui Xia. 2024.
\newblock \href {https://arxiv.org/abs/2304.04339} {Is chatgpt a good sentiment analyzer? a preliminary study}.
\newblock \emph{Preprint}, arXiv:2304.04339.

\bibitem[{Wang et~al.(2026)Wang, Cao, Xu, Ding, Gao, Mao, Qi, and Wang}]{wang_ordinal_2026}
Zhongyan Wang, Yuchen Cao, Shuo Xu, Zhanyi Ding, Yijun Gao, Qinkai Mao, Tongsong Qi, and Hailiang Wang. 2026.
\newblock \href {https://doi.org/10.3389/fdgth.2026.1836536} {Ordinal sentiment classification in cancer support forums: a controlled benchmark of machine learning and transformer models}.
\newblock \emph{Frontiers in Digital Health}, 8:1836536.

\bibitem[{Xie et~al.(2025)Xie, Zhang, Bao, Wang, Sun, Wang, and Xu}]{xie-etal-2025-comprehensive}
Guangyu Xie, Yice Zhang, Jianzhu Bao, Qianlong Wang, Yang Sun, Bingbing Wang, and Ruifeng Xu. 2025.
\newblock \href {https://doi.org/10.18653/v1/2025.emnlp-main.1122} {Comprehensive and efficient distillation for lightweight sentiment analysis models}.
\newblock In \emph{Proceedings of the 2025 Conference on Empirical Methods in Natural Language Processing}, pages 22070--22091, Suzhou, China. Association for Computational Linguistics.

\bibitem[{Yang et~al.(2021)Yang, Zha, Chen, Wang, and Katabi}]{yang_2021}
Yu-Zhe Yang, Kai-Wen Zha, Ying-Cong Chen, Hongya Wang, and Dina Katabi. 2021.
\newblock \href {https://api.semanticscholar.org/CorpusID:231951565} {Delving into deep imbalanced regression}.
\newblock In \emph{International Conference on Machine Learning}.

\bibitem[{Zhang et~al.(2024)Zhang, Deng, Liu, Pan, and Bing}]{zhang-etal-2024-sentiment}
Wenxuan Zhang, Yue Deng, Bing Liu, Sinno Pan, and Lidong Bing. 2024.
\newblock \href {https://doi.org/10.18653/v1/2024.findings-naacl.246} {Sentiment analysis in the era of large language models: A reality check}.
\newblock In \emph{Findings of the {A}ssociation for {C}omputational {L}inguistics: {NAACL} 2024}, pages 3881--3906, Mexico City, Mexico. Association for Computational Linguistics.

\bibitem[{Zhang et~al.(2025)Zhang, Xie, Lin, Bao, Wang, Zeng, and Xu}]{zhang-etal-2025-targeted}
Yice Zhang, Guangyu Xie, Jingjie Lin, Jianzhu Bao, Qianlong Wang, Xi~Zeng, and Ruifeng Xu. 2025.
\newblock \href {https://doi.org/10.18653/v1/2025.emnlp-main.1127} {Targeted distillation for sentiment analysis}.
\newblock In \emph{Proceedings of the 2025 Conference on Empirical Methods in Natural Language Processing}, pages 22158--22181, Suzhou, China. Association for Computational Linguistics.

\bibitem[{Öhman et~al.(2024)Öhman, Bizzoni, Moreira, and Nielbo}]{ohman_emotionarcs_2024}
Emily Öhman, Yuri Bizzoni, Pascale~Feldkamp Moreira, and Kristoffer~L Nielbo. 2024.
\newblock {EmotionArcs}: Emotion arcs for 9,000 literary texts.
\newblock In \emph{Proceedings of {LaTeCH}-{CLfL} 2024}, pages 51--66. Association for Computational Linguistics.

\end{thebibliography}

\appendix

\section{Teacher Model Data Preparation}
\label{app:annotation}

\subsection{Annotation Prompt}

The sentiment annotations were generated using
Gemma 4 31B through Ollama. Each sentence was
processed independently using the following prompt,
where \texttt{INPUT\_SENTENCE} was replaced with
the corresponding sentence from the corpus:

\begin{Verbatim}[
    fontsize=\scriptsize,
    breaklines=true,
    breakanywhere=true,
    breaksymbolleft={},
    breaksymbolright={},
    breaksymbolindentleft=0pt
]
You will perform the sentiment analysis of the sentence. Sentiment is the
emotional tone, attitude, or opinion expressed in text.

Evaluate the sentence you will get as an input in the scale from 1 to 10 where
1 = most negative and 10 = most positive sentiment.
Also, provide a brief explanation of your decision to score the sentiment in the sentence.

Output the results as json in the following format:

[
  {
    "Sentiment_score": your score from 1 to 10,
    "Explanation": your explanation for the score
  }
]

Here is the sentence: INPUT_SENTENCE
\end{Verbatim}

\subsection{Generation Parameters and Post-processing}

Inference was performed locally using Ollama with the model's default generation parameters in instant mode. No additional modifications to the default inference settings were applied.

To ensure consistent annotation output, the processing pipeline implemented a JSON parsing and validation procedure. Each sentence was allowed up to three generation attempts if the model failed to produce a valid, parsable JSON response.

However, Gemma 4 31B successfully generated valid JSON responses containing sentiment scores on the first attempt for all 10,889 sentences. Consequently, no additional generation attempts were required.

The resulting sentiment scores were extracted from the \texttt{Sentiment\_score} field and used as ordinal labels for subsequent BERT fine-tuning.

\section{List of BERT Base Models for Fine-Tuning}
\label{app:bert_models} 

Table~\ref{tab:bert_configurations} summarizes the six
pretrained BERT configurations considered in this study.
BERT-Large and BERT-Base follow the original architectures
introduced by \citet{devlin-etal-2019-bert}, while the
Medium, Small, Mini, and Tiny variants are compact models
introduced by \citet{turc2020wellread}.

All models are English uncased BERT encoders.
The configurations differ in the number of Transformer
layers, hidden dimensions, attention heads, and
trainable parameters. Parameter counts are approximate
and refer to the pretrained architectures.

\begin{table}[tb]
\centering
\small
\setlength{\tabcolsep}{3pt}
\begin{tabular}{@{}lrrrr@{}}
\hline
\textbf{Model} & \textbf{Layers} & \textbf{Hidden}
& \textbf{Heads} & \textbf{Params.} \\
\hline
BERT-Large  & 24 & 1024 & 16 & 340M \\
\multicolumn{5}{@{}l}{\texttt{google-bert/bert-large-uncased}} \\[2pt]
BERT-Base   & 12 & 768 & 12 & 110M \\
\multicolumn{5}{@{}l}{\texttt{google-bert/bert-base-uncased}} \\[2pt]
BERT-Medium & 8 & 512 & 8 & 41.7M \\
\multicolumn{5}{@{}l}{\texttt{google/bert\_uncased\_L-8\_H-512\_A-8}} \\[2pt]
BERT-Small  & 4 & 512 & 8 & 29.1M \\
\multicolumn{5}{@{}l}{\texttt{google/bert\_uncased\_L-4\_H-512\_A-8}} \\[2pt]
BERT-Mini   & 4 & 256 & 4 & 11.3M \\
\multicolumn{5}{@{}l}{\texttt{google/bert\_uncased\_L-4\_H-256\_A-4}} \\[2pt]
BERT-Tiny   & 2 & 128 & 2 & 4.4M \\
\multicolumn{5}{@{}l}{\texttt{google/bert\_uncased\_L-2\_H-128\_A-2}} \\
\hline
\end{tabular}
\caption{Architectural characteristics of the pretrained BERT
models. Each Hugging Face model ID appears below its model.
Parameter counts are approximate.}
\label{tab:bert_configurations}
\end{table}

\section{Fine-Tuning Training Parameters}
\label{app:training}

All BERT variants were fine-tuned using the Hugging Face Trainer framework with the same optimization hyperparameters. Training was configured for up to three epochs, with a fixed random seed of 1111. BERT-large and BERT-base used a maximum sequence length of 256 tokens, while the remaining models used 128 tokens.

\begin{table}[h]
\centering
\small
\begin{tabular}{ll}
\hline
\textbf{Hyperparameter} & \textbf{Value} \\
\hline
Learning rate & $2\times10^{-5}$ \\
Weight decay & 0.01 \\
Warmup ratio & 0.10 \\
Maximum epochs & 3 \\
Training batch size & 16 \\
Evaluation batch size & 32 \\
Gradient accumulation & 1 \\
Evaluation interval & 100 steps \\
Checkpoint interval & 100 steps \\
Logging interval & 100 steps \\
Checkpoint selection & Min. val. MAE \\
Maximum checkpoints & 6 \\
Random seed & 1111 \\
Mixed precision & FP16 (CUDA only) \\
Max. length (Large, Base) & 256 \\
Max. length (other models) & 128 \\
\hline
\end{tabular}
\caption{Training hyperparameters shared across
the ordinal BERT configurations.}
\label{tab:training_parameters}
\end{table}

Model evaluation, checkpoint saving, and logging were scheduled every 100 training steps. The best checkpoint for each model was selected according to the lowest validation mean absolute error (MAE). At most six checkpoints were retained, and FP16 mixed precision was enabled when CUDA was available. The training configuration is summarized in Table~\ref{tab:training_parameters}.

\subsection{LLM-Annotated Label Distribution}
\label{app:label_distribution}

The distribution of LLM-generated sentiment labels
across the complete corpus ($n=10,889$) is presented
in Table~\ref{tab:label_distribution}.
The annotations were unevenly distributed, with
score 5 accounting for 41.5\% of all sentences.
Overall, 57.9\% of annotations fell within the
near-neutral range (scores 4--6).

\begin{table}[tb]
\centering
\small
\begin{tabular}{rrrr}
\hline
\textbf{Score} & \textbf{Count} &
\textbf{Score} & \textbf{Count} \\
\hline
1 & 239  & 6  & 960 \\
2 & 800  & 7  & 637 \\
3 & 1436 & 8  & 923 \\
4 & 821  & 9  & 474 \\
5 & 4519 & 10 & 80  \\
\hline
\end{tabular}
\caption{Distribution of LLM-generated sentiment
labels across the complete annotated corpus
($n=10,889$), before the training, validation,
and test split.}
\label{tab:label_distribution}
\end{table}

\subsection{Evaluation Metrics and Checkpoint Selection}

During training, model performance was evaluated on the LLM-annotated validation set using MAE, root mean squared error (RMSE), accuracy, accuracy within one sentiment level, quadratic weighted Cohen's kappa (QWK),
and Spearman's rank correlation ($\rho$).

The checkpoint selection criterion was validation MAE, with lower values preferred. Following training, the models were also evaluated on the held-out test split of 1,089 LLM-annotated sentences. Table~\ref{tab:training_evaluation} reports the selected checkpoint steps, validation MAE, and test-set performance.

\begin{table}[tb]
\centering
\small
\setlength{\tabcolsep}{2.5pt}
\begin{tabular}{@{}lrrrr@{}}
\hline
\textbf{Model} & \textbf{Step} & \textbf{Val. MAE}
& \textbf{Test MAE} & \textbf{RMSE} \\
\hline
BERT-large  & 1600 & 0.847 & 0.838 & 1.196 \\
BERT-base   & 600  & 0.971 & 0.959 & 1.312 \\
BERT-medium & 1100 & 1.012 & 1.078 & 1.505 \\
BERT-small  & 1100 & 1.079 & 1.142 & 1.546 \\
BERT-mini   & 1600 & 1.219 & 1.278 & 1.633 \\
BERT-tiny   & 1600 & 1.354 & 1.353 & 1.895 \\
\hline
\end{tabular}
\vspace{3pt}
\begin{tabular}{@{}lrrrr@{}}
\hline
\textbf{Model} & \textbf{Acc.} & \textbf{Acc. $\pm$1}
& \textbf{QWK} & \textbf{$\rho$} \\
\hline
BERT-large  & 0.461 & 0.814 & 0.746 & 0.763 \\
BERT-base   & 0.401 & 0.771 & 0.680 & 0.711 \\
BERT-medium & 0.372 & 0.742 & 0.593 & 0.567 \\
BERT-small  & 0.360 & 0.699 & 0.529 & 0.532 \\
BERT-mini   & 0.264 & 0.661 & 0.445 & 0.479 \\
BERT-tiny   & 0.415 & 0.579 & 0.000 & 0.226 \\
\hline
\end{tabular}
\caption{Validation and test performance on LLM-annotated
sentences. Step identifies the selected checkpoint;
$\rho$ denotes Spearman's rank correlation.}
\label{tab:training_evaluation}
\end{table}

The models were subsequently evaluated against the independent human annotated Hemingway corpus, as reported in Section~\ref{sec:results}. These human annotations were not used for checkpoint selection.
\end{document}